\documentclass[letterpaper]{article} 
\usepackage{aaai24}  
\usepackage{times}  
\usepackage{helvet}  
\usepackage{courier}  
\usepackage[hyphens]{url}  
\usepackage{graphicx} 
\usepackage{natbib}  
\usepackage{caption} 
\usepackage{algorithm}
\usepackage{algorithmicx}

\usepackage{newfloat}
\usepackage{listings}
\DeclareCaptionStyle{ruled}{labelfont=normalfont,labelsep=colon,strut=off} 
\floatstyle{ruled}
\newfloat{listing}{tb}{lst}{}
\floatname{listing}{Listing}
\usepackage{supertabular}
\usepackage{colortbl}
\usepackage{bbding}
\usepackage[utf8]{inputenc} 
\usepackage[T1]{fontenc}    
\usepackage{microtype}      
\usepackage{xspace}
\usepackage{amsmath,amssymb,amsfonts,dsfont,pifont,bm,mathrsfs,mathtools,nicefrac}
\usepackage{algorithm,algpseudocode,listings}
\usepackage{booktabs,multirow,adjustbox,diagbox,threeparttable}
\usepackage{cleveref}  
\usepackage{lipsum}
\crefname{section}{Sec.}{Secs.}
\Crefname{section}{Section}{Sections}
\crefname{appendix}{Appendix}{Appendixes}
\crefname{table}{Tab.}{Tabs.}
\Crefname{table}{Table}{Tables}
\crefname{figure}{Fig.}{Figs.}
\Crefname{figure}{Figure}{Figures}
\crefname{equation}{Eq.}{Eqs.}
\Crefname{equation}{Equation}{Equations}
\newcommand{\method}{SemAIM\xspace}
\newcommand{\vm}[1]{\mbox{\boldmath $#1$}}
\newcommand{\supp}{\textit{Supplementary Material}\xspace}
\newcommand{\gc}[1]{}

\newlength\savewidth

\title{Semantic-Aware Autoregressive Image Modeling \\ for Visual Representation Learning}
\author{
    Kaiyou Song\textsuperscript{}\thanks{Corresponding author (songkaiyou@foxmail.com).}  \hspace{0.6mm}
    Shan Zhang  \hspace{0.6mm}
    Tong Wang  \hspace{0.6mm}\\[8pt]
}
\affiliations{
     Megvii Technology \\
     \{songkaiyou, zhangshan, wangtong\}@megvii.com
}

\usepackage{bibentry}

\begin{document}

\maketitle

\begin{abstract}

The development of autoregressive modeling (AM) in computer vision lags behind natural language processing (NLP) in self-supervised pre-training.
This is mainly caused by the challenge that images are not sequential signals and lack a natural order when applying autoregressive modeling.
In this study, inspired by human beings' way of grasping an image, i.e., focusing on the main object first, we present a semantic-aware autoregressive image modeling (\method) method to tackle this challenge.
The key insight of \method is to autoregressive model images from the semantic patches to the less semantic patches.
To this end, we first calculate a semantic-aware permutation of patches according to their feature similarities and then perform the autoregression procedure based on the permutation.
In addition, considering that the raw pixels of patches are low-level signals and are not ideal prediction targets for learning high-level semantic representation, we also explore utilizing the patch features as the prediction targets.
Extensive experiments are conducted on a broad range of downstream tasks, including image classification, object detection, and instance/semantic segmentation, to evaluate the performance of \method.
The results demonstrate \method achieves state-of-the-art performance compared with other self-supervised methods.
Specifically, with ViT-B, \method achieves 84.1\% top-1 accuracy for fine-tuning on ImageNet, 51.3\% AP and 45.4\% AP for object detection and instance segmentation on COCO, which outperforms the vanilla MAE by 0.5\%, 1.0\%, and 0.5\%, respectively.
Code is available at \url{https://github.com/skyoux/SemAIM}.

\end{abstract}

\section{Introduction}\label{sec:intro}

With the rapid development of masked language modeling (MLM) (e.g., BERT~\cite{bert2018}) and autoregressive language modeling (ALM) (e.g., GPT~\cite{gpt1_2018, gpt2_2019, gpt3_2020}), self-supervised pre-training has achieved impressive performance in learning extensible representations in the field of natural language processing (NLP).
Recently, inspired by the masking mechanism in MLM, masked image modeling (MIM)~\cite{beit2021, mae2022, ibot2021, simmim2022} has been proposed and rapidly improved in the computer vision community.
The key to applying MIM is to use a high mask ratio on images to reduce spatial redundancy.
MIM achieves a better performance of self-supervised learning (SSL) in many downstream tasks~\cite{imagenet, coco, ade20k} compared with other alternatives~\cite{moco_2020, simclr_2020, mocov3_2021, dino_2021}.
By contrast, the development of autoregressive image modeling (AIM) lags behind MIM in computer vision due to the significant difference between language and vision.

As discussed in~\cite{mae2022}, languages are human-generated signals with high-level semantics and dense information.
They are sequential signals and provide a natural order for applying autoregressive modeling.
For example, the "left-to-right" ALM used in GPT~\cite{gpt1_2018, gpt2_2019, gpt3_2020} shows a strong language generation and understanding ability.
On the contrary, images are natural signals with heavy spatial redundancy.
They are not sequential signals and lack a natural order for applying autoregressive modeling.
Therefore, autoregressive image modeling is not an effortless way compared to masked image modeling.
Several studies~\cite{igpt2020, randsac2022, saim2022} attempt to conduct autoregressive image modeling for self-supervised learning.
They propose to use raster order image pixels~\cite{igpt2020} and stochastic order image patches~\cite{randsac2022, saim2022} for autoregressive modeling.
However, we argue that raster and stochastic orders are not ideal for visual representation learning since they are inconsistent with the human visual mechanisms of grasping an image.
According to the attention mechanism of the human visual system, human beings always selectively attend to the most informative parts of visual stimuli~\cite{1972_humanattention, 2006_humanattention}.
Specifically, they first focus on the main object or the object they are interested in, then focus on other contents in images, such as the background and other objects.
This attention mechanism makes human beings grasp an image quickly.

Can we apply autoregressive image modeling by mimicking human visual mechanisms of grasping an image (i.e., focusing on the main object first).
In this study, we design a semantic-aware autoregressive image modeling (\method) method to answer this question.
The core concept of \method is to model images from the semantic patches to the less semantic patches autoregressively.
To achieve this goal, we first calculate a semantic-aware order of patches according to their feature similarities and then perform the autoregression procedure based on the permutation.
Furthermore, considering that the raw pixels of patches are low-level signals and are not ideal prediction targets for learning high-level semantic representation, we also explore utilizing the patch features as the prediction targets.
We conducted extensive experiments on a broad range of downstream tasks, including image classification, object detection, and instance/semantic segmentation, to evaluate the performance of \method.
The experimental results show \method achieves state-of-the-art performance compared with other self-supervised methods, especially in dense prediction tasks.
Specifically, with ViT-B, \method achieves 84.1\% top-1 accuracy for fine-tuning on ImageNet, 51.3\% AP on COCO for object detection, and 45.4\% AP on COCO for instance segmentation, which outperforms the vanilla MAE by 0.5\%, 1.0\%, and 0.5\%, respectively.
This study empirically demonstrates that \method is more appropriate for autoregressive image modeling and is helpful for learning semantic visual representation.
\section{Related work}\label{sec:related}

\textbf{Self-Supervised Learning.} 
Self-supervised learning aims to learn scalable visual representations without any human annotations.
The key to self-supervised learning is how to design an effective pretext task to learn semantic representations.
In the field of computer vision, early studies designed various pretext tasks, including image inpainting~\cite{pathak2016context}, colorization~\cite{zhang2016colorful}, jigsaw puzzle~\cite{noroozi2016unsupervised}, counting~\cite{noroozi2017representation}, and rotation prediction~\cite{komodakis2018unsupervised}.
Though with inferior performance, these studies laid the foundation for the development of this field.
After that, contrastive learning~\cite{moco_2020, simclr_2020, swav_2020, simsiam_2021, mocov3_2021, dino_2021, hcsc_2022, mokd_2023, scfs_2023}, as a type of instance discriminative~\cite{insdis_2018} method, is heavily studied and has shown remarkable progress in recent years, which aims at pulling different augmented versions of the same image closer while pushing diverse images far from each other. 
However, contrastive learning needs curated data and carefully-designed data augmentation techniques for pre-training~\cite{simclr_2020, dino_2021}.

\noindent\textbf{Masked Image Modeling.}
Inspired by mask language modeling~\cite{bert2018} in NLP, masked image modeling~\cite{beit2021, mae2022} has been proposed for visual pre-training in the recent two years.
Many studies are proposed, which mainly focus on improving the performance of masked image modeling from two aspects, i.e., prediction targets and masking strategy.
For prediction targets, various contents are explored, including raw RGB pixels~\cite{mae2022, simmim2022, cae2022, hpm2023}, discrete tokens~\cite{beit2021, peco2022}, HoG features~\cite{maskfeat2022}, features extracted from a momentum model~\cite{ibot2021, bootmae2022}), and features extracted from pre-trained models~\cite{semmae2022, milan2022}.
For the masking strategy, random masking~\cite{mae2022, simmim2022} and block-wise masking strategy~\cite{beit2021, ibot2021} are adopted in many methods.
While some studies~\cite{attmask2022, semmae2022, hpm2023} explore adaptive or learnable masking strategies to force the model to learn semantic visual representations.

\noindent\textbf{Autoregressive Modeling.}
Autoregressive language modeling (ALM) proposed in GPT~\cite{gpt1_2018, gpt2_2019, gpt3_2020} has shown significant success in learning general representations in the field of NLP.
ALM in GPT predicts the next possible word based on the preceding word in left-to-right order.
Besides left-to-right prediction, permuted ALM~\cite{yang2019xlnet, song2020mpnet} aims to learn contextual information by maximizing the expected logarithmic likelihood of all possible permutations of sequences.
While UniLM~\cite{dong2019unified} adopts both left-to-right and right-to-left predictions to boost the performance of pre-trained language models.

While in the field of computer vision, autoregressive image modeling (AIM) is employed for several tasks, such as image generation~\cite{van2016conditional, ramesh2021zero}, object detection~\cite{chen2021pix2seq}, and representation learning~\cite{cpc2018, igpt2020, randsac2022, saim2022}.
For the representation learning task, early study CPC~\cite{cpc2018} uses autoregressive models to learn representations by predicting the future in latent space.
iGPT~\cite{igpt2020} directly apples GPT~\cite{gpt1_2018} on images by autoregressive modeling raw pixels in raster order. 
Serializing images into pixels is not an ideal strategy, which greatly limits the efficiency and representation learning performance of iGPT.
After that, with the development of vision transformer~\cite{transformer2017, vit2020}, which serializes images into patches, recent studies~\cite{randsac2022, saim2022} conduct autoregressive modeling based on image patches.
RandSAC~\cite{randsac2022} groups patches into hierarchically arranged segments and performs autoregressive prediction on these segments in stochastic order.
SAIM~\cite{saim2022} directly applies autoregressive prediction on patches in stochastic order.
However, we argue that raster or stochastic orders are not ideal for visual representation learning since they are inconsistent with the human visual mechanisms of grasping an image, i.e., focusing on the main object first.
In this study, we present the semantic-aware autoregressive image modeling (\method) to tackle this problem.

\section{Method}\label{sec:method}

\begin{figure*}[t]
\centering
\includegraphics[width=0.8\linewidth]{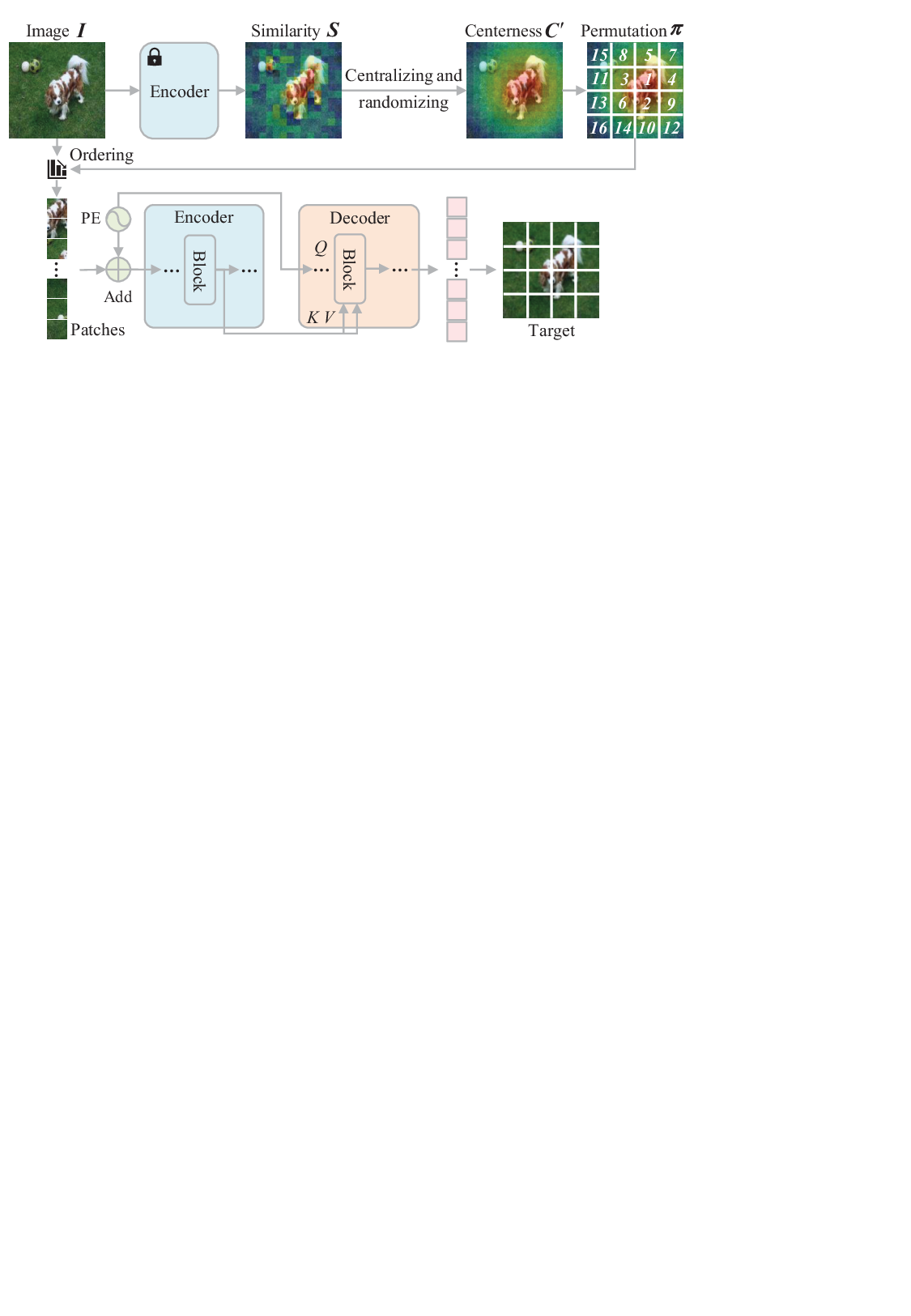}
\caption{
\textbf{Illustration of \method.}
Given an input image $\vm{I}$, we first calculate its similarity map $\vm{S}$ and generate a semantic-aware permutation $\vm{\pi}$.
Then, we employ a parallel encoder-decoder for autoregressive modeling according to the permutation.
``PE" denotes the position embedding, and ``add" denotes element-wise addition.
Note that the centerness in \cref{equ_add_randomness} is processed by ${\vm{C}'}=1-{\rm{softmax}}(\vm{C})$ for visualization.
At fine-tuning stage, the encoder is applied for downstream tasks.
}
\label{fig:pipeline}
\end{figure*}

\subsection{Preliminary: Autoregressive Modeling}

Autoregressive modeling aims to learn a good visual representation from an unlabeled dataset by modeling its distribution.
Specifically, given an unlabeled dataset ${\cal D}$ consisting of high dimensional data $\vm{x} = \left[ {{x_1},{x_2}, \ldots ,{x_N}} \right]$, a permutation $\vm{\pi}$ of the set $[1,N]$ can be picked, and ${\vm{\pi}}_i$ and ${\vm{\pi}}_{<i}$ denote the $i$-th element and the first 1-$i$ elements of the permutation.
Autoregressive modeling learns the data distribution by maximizing the likelihood function:
\begin{equation}
{\cal L} =  - \mathop{\mathbb{E}} \limits_{x \sim {\cal D}} \sum\limits_{i = 1}^N {\log {p_\theta }} \left( {{x_{{\vm{\pi} _i}}}|{x_{{\vm{\pi} _{ < i}}}}} \right)
\label{equ_loss_ar}
\end{equation}
where $\theta$ is the parameters of the autoregressive model.
When working with images, an image is reshaped into a sequence of pixels~\cite{igpt2020} or patches~\cite{randsac2022,saim2022}, and the permutation is generated by a fixed raster order~\cite{igpt2020} or a stochastic order~\cite{randsac2022,saim2022}.
However, as we have analyzed before, such orders are inconsistent with the human visual understanding, i.e., focusing on the semantic object first.

\subsection{Semantic-guided Autoregressive Image Modeling}\label{sec:semaim}

In this study, we present the semantic-aware autoregressive image modeling (\method) to overcome the limitations of existing autoregressive image modeling methods.
\cref{fig:pipeline} illustrates our proposed \method.
In \method, we first calculate its similarity map first and generate semantic-aware permutation of its patches.
Then, we employ a parallel encoder-decoder for autoregressive modeling according to the generated permutation.

Given an image $\vm{I} \in \mathbb{R}^{H \times W \times C}$, it is first reshaped into a sequence of patches ${\vm{I}}_p \in \mathbb{R}^{N \times (P^2 C)}$, where $(H, W)$ indicates the spatial resolution, $C$ is the number of channels, $P$ is the patch size, and $N=HW/P^2$ is the number of patches.
A linear projection is then applied to ${\vm{I}}_p$, mapping it to $D$ dimensions to get patch embeddings $\vm{x}_p \in \mathbb{R}^{N \times D}$.
A \texttt{[CLS]} token $\vm{x}_{\mathrm{cls}} \in \mathbb{R}^{D}$ is used to aggregate the information.
2D sin-cos position embeddings $\vm{p} \in \mathbb{R}^{(N+1) \times D}$ are added to the patch embeddings to retain positional information.
Thus, the initialized sequence $\vm{x} = [\vm{x}_{\mathrm{cls}}; \vm{x}_p] \oplus \vm{p}$ can be obtained.
Where $\oplus$ denotes element-wise addition.

\subsubsection{Semantic-aware Permutation Generation}

To generate semantic-aware order, we need to locate semantic regions first.
In this study, we found that the similarities among the \texttt{[CLS]} token and patch tokens from the deep layers of the pre-trained encoder can locate the semantic regions of input images.
Therefore, we first generate semantic-aware permutation according to the similarity map.

Feed the embedded tokens $\vm{x}$ into the frozen encoder, we can get the output tokens $\vm{z} = \left[ {{{\vm{z}}_{cls}},\vm{z}_p} \right]$ from the last blocks of the encoder.
Then, the similarities among the \texttt{[CLS]} token ${\vm{z}}_{cls}$ and patch tokens $\vm{z}$ can be calculated:
\begin{equation}
{\vm{S}_i} = \frac{{\exp \left( {\cos ({\vm{z}_{cls}},{\vm{z}_i})} \right)}}{{\sum\nolimits_{j = 1}^N {\exp \left( {\cos ({\vm{z}_{cls}},{\vm{z}_j})} \right)} }}
\label{equ_similarity}
\end{equation}
where $\vm{S} \in \mathbb{R}^N$ denotes the similarity matrix, and $\cos \left( {} \right)$ denotes cosine similarity between two vectors.
Concretely, $\vm{S}_i$ means the similarity between the \texttt{[CLS]} token and the $i$-th patch token.
We can reshape $\vm{S}$ into two-dimension map $\vm{S} \in \mathbb{R}^{{H'} \times {W'}}$, where $H' = H/P$, $W' = W/P$.
The similarity map $\vm{S}$ highlights the semantic regions in the image since the \texttt{[CLS]} token aggregates the global information.

In this study, we found that it is not ideal to use the similarity map directly for autoregression permutation due to two reasons.
First, the similarity map is not an accurate semantic segmentation map, which is noisy and not a smooth permutation.
Second, using the similarity map directly will significantly decrease the diversity of autoregression.
Therefore, based on the similarity map, we further design a center-to-outward permutation motivated by human visual mechanisms of grasping an image.
We can get the semantic region center according to the similarity map $\vm{S}$:
\begin{equation}
{c_y},{c_x} = \arg \max \left( {{\vm{S}_{ij}}} \right)
\label{equ_center}
\end{equation}
Note that we adopt a $3\times3$ mean filtering operation on $\vm{S}$ to alleviate the influence of noise.
The patch with index ${c_y},{c_x}$ is termed as the center patch.
Then, the distances between patches and center are calculated:
\begin{equation}
{\vm{D}_{ij}} = \sqrt {{{\left( {{c_y} - i} \right)}^2} + {{\left( {{c_x} - j} \right)}^2}}
\label{equ_sort}
\end{equation}
Not that patches located on the same radius have the same distance and show a similar level of semantics.
To increase the diversity for autoregression, we randomly generate a vector ${\vm{R}_{ij}} = {\rm{U}}\left( {0,1} \right)$ (where $\rm{U}$ denotes the uniform distribution), and add it to the distance $\vm{D}$ and obtain the centerness $\vm{C}$:
\begin{equation}
{\vm{C}_{ij}} = {\vm{D}_{ij}} + \lambda{\vm{R}_{ij}}
\label{equ_add_randomness}
\end{equation}
where $\lambda=0.01$ is set to avoid the randomness change the order for patches with different distances.
The centerness $\vm{C}$ can be reshaped to one dimension $\vm{C} \in \mathbb{R}^N$.
Finally, the autoregression permutation can be calculated:
\begin{equation}
\vm{\pi}  = \arg {\rm{sort}}(\vm{C})
\label{equ_permutation}
\end{equation}
Thus, we generated the semantic-aware autoregressive permutation for image patches.

\subsubsection{Autoregressive Modeling}

Based on the generated autoregressive permutation $\vm{\pi}$, the autoregressive modeling procedure can be conducted.
Following previous work~\cite{saim2022}, we design a parallel encoder-decoder architecture to perform autoregressive modeling.
During pre-training, the encoder focuses on learning contextual information, and the decoder focuses on predicting the given target of the original image from the latent representation.
During fine-tuning, only the encoder will be reserved and fine-tuned for downstream tasks.

\noindent\textbf{Encoder.}
The encoder learns contextual information with masked self-attention.
It has the same structure as the Vision Transformer~\cite{vit2020}, which consists of $L$ layers of self-attention blocks.
We apply a mask to the self-attention blocks to make the current token see only the preceding tokens in the permutation $\vm{\pi}$.
Specifically, the mask is generated as follows:
\begin{equation}
{\vm{M}_{ij}} = \left\{ \begin{array}{l}
\!0, \quad {\vm{C}_i} < {\vm{C}_j}\\
\!1, \quad {\vm{C}_i} \ge {\vm{C}_j}
\end{array} \right.
\label{equ_encoder_mask}
\end{equation}
where $\vm{M}\!\in\!{\mathbb{R}^{N \times N}}$, $i, j$ are the coordinate of attention matrix.
$\vm{M}_{ij}=1$ represents the $i$-th token have access to the $i$-th token, and vice verse.
We define the output of $l$-th encoder layer as $\vm{h}_i^{\left( l \right)}$, where $i$ is the token index.
And the initialized sequence $\vm{x}$ is the input of the first encoder layer, i.e. $\vm{h}_i^{\left( 0 \right)}=\vm{x}_i$.
Omitting the layer norm, MLP, and residual connection for simplification, the forward process of the encoder can be described as follows:
\begin{equation}
\vm{h}_{{\vm{\pi}}_t}^{(l)}={\rm Attention}({\rm Q}=\vm{h}_{{\vm{\pi}}_i}^{(l-1)},{\rm KV}=\vm{h}_{\boldsymbol {\vm{\pi}}_{\le i}}^{(l-1)};\theta_{e}^{(m)})
\label{equ_encoder_forward}
\end{equation}
where $1\le l \le L$, $\theta_{e}^{(l)}$ is the parameters of the $l$-th encoder layer.
Note that the masking strategy is implemented by adding a minus infinity value to the self-attention score where $\vm{M}_{ij}=0$.
Thus, during encoding, the current token can only see the preceding tokens.

\noindent\textbf{Decoder.}
The decoder consists of $L'$ layers of cross-attention blocks and an MLP head.
The blocks decode the input signals from the latent representation and the MLP head projects the signal to the dimension of the given target of the original image.
A mask is applied to the cross-attention blocks:
\begin{equation}
{\vm{M}'_{ij}} = \left\{ \begin{array}{l}
\!0, \quad {\vm{C}_i} \le {\vm{C}_j}\\
\!1, \quad {\vm{C}_i} > {\vm{C}_j}
\end{array} \right.
\label{equ_decoder_mask}
\end{equation}
Compared with the encoder mask in \cref{equ_encoder_mask}, the current token can see the preceding tokens and itself.
We can define the output of the $l$-th decoder layer as $\vm{g}_i^{(l)}$.
And the position embeddings $\vm{p}$ are used as the input of the first decoder layer, i.e., $\vm{g}_i^{(0)}=\vm{p}_i$.
The forward process of the decoder blocks can be formulated as follows:
\begin{equation}
\vm{g}_{{\vm{\pi}}_i}^{(l)}={\rm Attention}({\rm Q}=\vm{g}_{{\vm{\pi}}_i}^{(l-1)},{\rm KV}=\vm{h}_{\boldsymbol {\vm{\pi}}_{< i}}^{(l-1)};\theta_{d}^{(m)})
\label{equ_decoder_forward}
\end{equation}
where $1\le l \le L'$, $\theta_{d}^{(l)}$ is the parameters of the $l$-th decoder layer.
Finally, the MLP head projects the output of the decoder blocks to the target space:
\begin{equation}
\vm{g}_{{\vm{\pi}}_i}={\rm MLP}(\vm{g}_{{\vm{\pi}}_i}^{(L')};\theta_{h})
\label{equ_head_forward}
\end{equation}
where $\theta_{h}$ is the parameters of the MLP head.

\noindent\textbf{Prediction Targets.}
The autoregression model is trained by minimizing the mean squared error between the output prediction $\vm{g}_{{\vm{\pi}}_i}$ and given targets ${\hat{\vm{g}}}_{{\vm{\pi}}_i}$:
\begin{equation}
\mathcal L=\mathop {\mathbb E} \limits_{\boldsymbol x \sim \mathcal D} ~\sum_{i=1}^N || \vm{g}_{{\vm{\pi}}_i}-{\hat{\vm{g}}}_{{\vm{\pi}}_i} ||^2
\label{equ_l2_loss}
\end{equation}
In previous autoregression image modeling~\cite{saim2022}, the raw pixels of patches are employed as the prediction targets ${\hat{\vm{g}}}$.
In this study, considering that the raw pixels of patches are low-level signals and are not ideal prediction targets for learning high-level semantic representation, we also explore utilizing the patch features as the prediction targets.
Specifically, we also use the features extracted from the pre-trained ViT-B model of DINO~\cite{dino_2021} and CLIP~\cite{clip_2021} as the prediction targets.


\begin{table*}[t]
\centering
\begin{tabular}{c|lll|llllll|l}
\toprule
\multirow{2}{*}{order} & \multicolumn{3}{c|}{ImageNet} & \multicolumn{3}{c}{COCO detection} & \multicolumn{3}{c|}{COCO segmentation} & ADE20k \\
& fine-tune & \gc{linear} & \gc{$k$-NN} & AP$^{\text{b}}$ & \gc{AP$^{\text{b}}_{\text{50}}$} & \gc{AP$^{\text{b}}_{\text{75}}$} & AP$^{\text{m}}$ & \gc{AP$^{\text{m}}_{\text{50}}$} & \gc{AP$^{\text{m}}_{\text{75}}$} & mIoU \\
\midrule
raster & 82.9 & \gc{27.8} & \gc{13.7} & 42.0 & \gc{62.5} & \gc{46.2} & 37.9 & \gc{60.0} & \gc{40.5} & 42.2 \\
stochastic & 83.0 & \gc{55.5} & \gc{29.1} & 42.1 & \gc{62.8} & \gc{46.4} & 38.0 & \gc{59.9} & \gc{40.8} & 42.4 \\
similarity & 82.4 & \gc{42.2} & \gc{17.1} & 38.9 & \gc{59.2} & \gc{42.5} & 35.4 & \gc{56.2} & \gc{37.8} & 41.6 \\
%
\textbf{semantic-aware} & \textbf{83.5} & \gc{\textbf{55.8}} & \gc{\textbf{35.3}} & \textbf{43.3} & \gc{\textbf{64.0}} & \gc{\textbf{47.8}} & \textbf{39.1} & \gc{\textbf{61.1}} & \gc{\textbf{42.0}} & \textbf{43.9} \\
\bottomrule
\end{tabular}
\caption{
\textbf{Ablation study of autoregression order.}
The RGB value is adopted as the prediction target, and the decoder depth is 12.
Note that ``similarity'' denotes the similarity is directly used as autoregression order.
}
\label{tab:ablation_order}
\end{table*}

\begin{figure*}[t]
\centering
\includegraphics[width=0.98\linewidth]{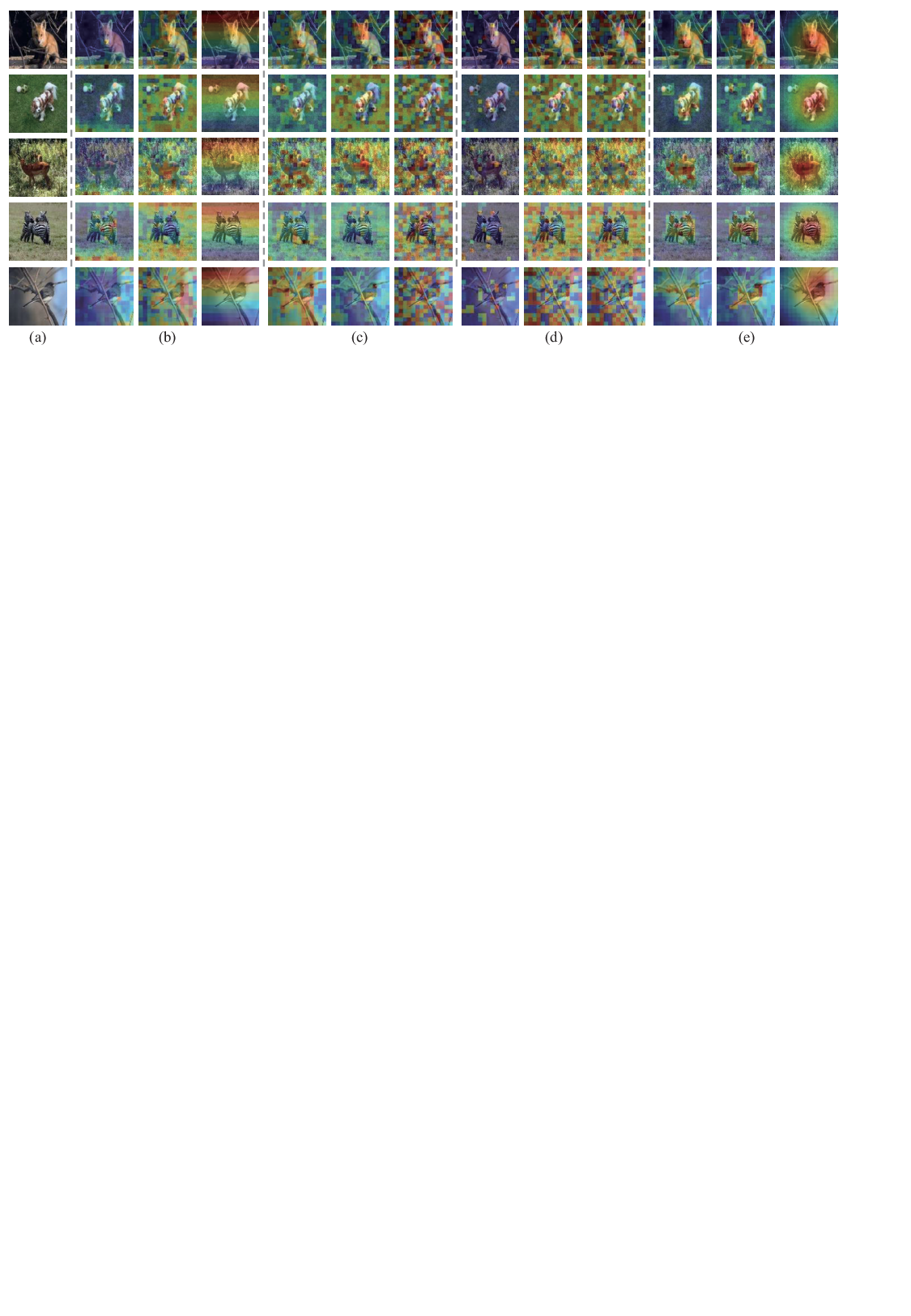}
\caption{
Visualization of different autoregression orders.
(a) input images, (b) raster order used in iGPT~\cite{igpt2020}, (c) stochastic order used in SAIM~\cite{saim2022}, (d) similarity order (the similarity map $S$ is also directly used as the autoregression order), and (e) semantic-aware order used in \method. In (b)(c)(d)(e), the first column shows the self-attention maps from the last block, the second column shows similarity maps $S$ from the last block, and the last column shows the corresponding autoregression orders (more warm-colored patches are predicted first).
}
\label{fig_visual}
\end{figure*}

\section{Experiments}\label{sec:exp}

\subsection{Settings}

\textbf{Pre-training.}
All self-supervised pre-training is performed on the ImageNet~\cite{imagenet} training set with a resolution of 224$\times$224.
In default settings, we take ViT-B/16~\cite{vit2020} as the default backbone and pre-train models with a 2048 batch size for 200 epochs.
The decoder is a stack of cross-attention Transformer~\cite{vit2020} blocks and has 12 blocks.
Details can be found in \supp.

\noindent\textbf{ImageNet classification.}
After pre-training, we perform end-to-end fine-tuning, linear probing, and $k$-NN classification on ImageNet~\cite{imagenet} to evaluate our \method.
For fine-tuning, 100 epochs with a 1024 batch size are performed following common practices~\cite{mae2022, beit2021} by default.
For linear probing, 90 epochs with a 4096 batch size are performed following common practices~\cite{mae2022, saim2022}.
The implementation of $k$-NN classification is based on DINO~\cite{dino_2021}.
We report top-1 accuracy of a single 224$\times$224 resolution on the ImageNet~\cite{imagenet} validation set.

\noindent\textbf{COCO object detection and instance segmentation.}
Following previous methods~\cite{mae2022, beit2021}, the Mask R-CNN~\cite{maskrcnn2017} with FPN~\cite{fpn2017} is adopted as the detector.
We conduct end-to-end fine-tuning on COCO~\cite{coco} with 1024$\times$1024 resolution.
We train 12.5 epochs with a 16 batch size for ablations and 100 epochs with a 64 batch size for fair comparison with other methods.
AP$^{\text{b}}$ and AP$^{\text{m}}$ are reported for object detection and instance segmentation, respectively.
Our implementation is based on detectron2~\cite{detectron2} and ViTDet~\cite{vitdet2022}.

\noindent\textbf{ADE20k semantic segmentation.}
Following previous works~\cite{mae2022, beit2021}, UperNet~\cite{upernet2018} is adopted as the decoder.
we perform end-to-end fine-tuning on ADE20k~\cite{ade20k} for 160k iterations with 512$\times$512 resolution and a 16 batch size.
mIoU~\cite{everingham2009pascal} is used for evaluation.
Our implementation is based on mmsegmentation~\cite{mmseg2020}.

\begin{table*}[t]
\centering
\begin{tabular}{c|lll|llllll|l}
\toprule
\multirow{2}{*}{targets} & \multicolumn{3}{c|}{ImageNet} & \multicolumn{3}{c}{COCO detection} & \multicolumn{3}{c|}{COCO segmentation} & ADE20k \\
& fine-tune & \gc{linear} & \gc{$k$-NN} & AP$^{\text{b}}$ & \gc{AP$^{\text{b}}_{\text{50}}$} & \gc{AP$^{\text{b}}_{\text{75}}$} & AP$^{\text{m}}$ & \gc{AP$^{\text{m}}_{\text{50}}$} & \gc{AP$^{\text{m}}_{\text{75}}$} & mIoU \\
\midrule
\textbf{RGB} & 83.5 & \gc{55.8} & \gc{35.3} & 43.3 & \gc{\textbf{64.0}} & \gc{47.8} & 39.1 & \gc{61.1} & \gc{42.0} & 43.9 \\
DINO & 84.4 & \gc{60.4} & \gc{54.1} & \textbf{47.8} & \gc{69.8} & \gc{\textbf{52.5}} & \textbf{42.3} & \gc{66.3} & \gc{\textbf{45.3}} & 48.3 \\
CLIP & \textbf{84.6} & \gc{\textbf{70.3}} & \gc{\textbf{56.6}} & 47.6 & \gc{\textbf{70.4}} & \gc{52.1} & 42.1 & \gc{\textbf{66.6}} & \gc{44.9} & \textbf{51.1} \\
\bottomrule
\end{tabular}
\caption{
\textbf{Ablation study of prediction targets.}
The semantic-aware order is adopted, and the decoder depth is 12.
}
\label{tab:ablation_targets}
\end{table*}

\begin{table*}[t]
\centering
\begin{tabular}{c|lll|llllll|l}
\toprule
\multirow{2}{*}{$d$} & \multicolumn{3}{c|}{ImageNet} & \multicolumn{3}{c}{COCO detection} & \multicolumn{3}{c|}{COCO segmentation} & ADE20k \\
& fine-tune & \gc{linear} & \gc{$k$-NN} & AP$^{\text{b}}$ & \gc{AP$^{\text{b}}_{\text{50}}$} & \gc{AP$^{\text{b}}_{\text{75}}$} & AP$^{\text{m}}$ & \gc{AP$^{\text{m}}_{\text{50}}$} & \gc{AP$^{\text{m}}_{\text{75}}$} & mIoU \\
\midrule
1 & 82.1 & \gc{50.2} & \gc{22.8} & 42.0 & \gc{62.4} & \gc{46.2} & 38.2 & \gc{59.5} & \gc{41.1} & 42.3 \\
6 & 83.3 & \gc{48.4} & \gc{25.1} & 43.1 & \gc{63.9} & \gc{47.5} & 39.0 & \gc{60.8} & \gc{41.8} & 43.3 \\
%
\textbf{12} & \textbf{83.5} & \gc{\textbf{55.8}} & \gc{\textbf{35.3}} & \textbf{43.3} & \gc{\textbf{64.0}} & \gc{\textbf{47.8}} & \textbf{39.1} & \gc{\textbf{61.1}} & \gc{\textbf{42.0}} & \textbf{43.9} \\
\bottomrule
\end{tabular}
\caption{
\textbf{Ablation study of decoder depth.}
The semantic-aware order is adopted, and the RGB value is adopted as the prediction target.
}
\label{tab:ablation_depth}
\end{table*}

\subsection{Ablation Study}

In this part, we conduct ablation studies to analyze the influence of each part of \method.
Specifically, we analyze the influence of autoregression order, prediction targets, and decoder depth in \cref{tab:ablation_order}, \cref{tab:ablation_targets}, and \cref{tab:ablation_depth}, respectively.
We take the ViT-B/16~\cite{vit2020} pre-trained with 200 epochs on ImageNet~\cite{imagenet} as the backbone and report the results on ImageNet~\cite{imagenet} classification, COCO~\cite{coco} object detection and instance segmentation, and ADE20k~\cite{ade20k} semantic segmentation.
In these experiments, 100 epochs of fine-tuning on ImageNet~\cite{imagenet}, 12.5 epochs of fine-tuning on COCO~\cite{coco}, and 160k iterations of fine-tuning on ADE20k~\cite{ade20k} are performed.
The default settings of our \method are shown in bold in the tables.

\noindent\textbf{Influence of autoregression order.}
In this experiment, we analyze the influence of autoregression order for representation learning.
The raster and stochastic order adopted in iGPT~\cite{igpt2020} and SAIM~\cite{saim2022} are compared with the semantic-aware order adopted in \method.
In addition, the similarity map $\vm{S}$ is also directly used as the autoregression order (patch with larger similarity is predicted first), which is denoted as ``similarity''.
As shown in \cref{tab:ablation_order}, the semantic-aware order proposed in this study significantly outperforms the raster and stochastic order, which verifies that semantic-aware prediction is more suitable for autoregression image modeling.
Further, the ``similarity'' order achieves the worst performance, which indicates that it is noisy and unstable to utilize the similarity map for autoregression order directly.

In addition, we visualize the self-attention maps, and the similarity maps $\vm{S}$, and the corresponding autoregression orders of each method in \cref{fig_visual}.
The self-attention maps and the similarity maps of the semantic-aware order used in \method locate on semantic regions more accurately than other methods.
This indicates that \method can learn more semantic representations.

\noindent\textbf{Influence of prediction targets.}
In this experiment, we analyze the influence of prediction targets.
Apart from the RGB value, we also utilize the feature of the pre-trained ViT-B model from DINO~\cite{dino_2021} and CLIP~\cite{clip_2021} as the prediction targets.
As shown in \cref{tab:ablation_targets}, using features as prediction targets perform better than RGB value, which indicates that utilizing high-level features as prediction targets is more beneficial for learning high-level semantic representation.

\noindent\textbf{Influence of decoder depth.}
In this experiment, we analyze the influence of the decoder depth.
\cref{tab:ablation_depth} varies the decoder depth $L'$ (number of Transformer blocks).
When the decoder depth is less than the encoder depth, i.e., $L'<L$, we uniformly choose $L'$ layer of the encoder to connect with the decoder during the parallel autoregression modeling procedure.
As can be seen, a sufficiently deep decoder is essential for the performance of \method.
This result demonstrates that autoregression image modeling needs a deeper decoder than masked image modeling~\cite{mae2022}.

\subsection{Comparisons with Other Methods}

The proposed \method is compared with a wide range of self-supervised counterparts on ImageNet~\cite{imagenet} classification, COCO~\cite{coco} object detection and instance segmentation, and ADE20k~\cite{ade20k} semantic segmentation.

We compare our \method with a wide range of self-supervised methods, including contrastive learning methods~\cite{dino_2021,mocov3_2021}, masked image modeling methods~\cite{beit2021,mae2022,simmim2022,semmae2022,localmim2023,hpm2023}, contrastive learning and masked image modeling combinations~\cite{ibot2021,bootmae2022}, and autoregression image modeling~\cite{igpt2020,randsac2022,saim2022}.
All methods are pre-trained with the same resolution, \textit{i.e.}, 224$\times$224 on ImageNet-1K~\cite{imagenet}.

\noindent\textbf{ImageNet classification.}
We compare the proposed \method with state-of-the-art alternatives on the ImageNet-1K~\cite{imagenet}.
The results are shown in \cref{tab:in1k}.
Notably, with only 400 epochs pre-training, \method achieves 83.8\% using ViT-B/16 as the backbone, surpassing MAE~\cite{mae2022} pre-trained for 1600 epochs by +0.2\%.
This empirical evidence demonstrates that enhancing the semantic awareness of ViTs during autoregression modeling brings better visual representations.
With 800 epochs of pre-training, \method outperforms its baseline SAIM~\cite{saim2022} by 0.2\% using ViT-B/16 as the backbone.
It achieves competitive results compared with the state-of-the-art.
Specifically, it achieves 84.1\% and 85.8\% using ViT-B/16 and ViT-L/16, respectively.
Furthermore, \method reaches state-of-the-art results, 85.3\% and 86.5\% using ViT-B/16 and ViT-L/16, respectively, by using CLIP~\cite{clip_2021} feature as predict targets.

\begin{table}[t]
\centering
\setlength{\tabcolsep}{3.5pt}
\begin{tabular}{lcll}
\toprule
method & ep. & ViT-B & ViT-L \\
\midrule
\multicolumn{4}{l}{\textit{Contrastive Learning}} \\
DINO~\cite{dino_2021} & 1600 & 82.8 & - \\
MoCo v3~\cite{mocov3_2021} & 600 & 83.2 & 84.1 \\
\midrule
\multicolumn{4}{l}{\textit{Masked Image Modeling}} \\
BEiT~\cite{beit2021} & 800 & 83.2 & 85.2 \\
MAE~\cite{mae2022} & 1600 & 83.6 & 85.9 \\
SimMIM~\cite{simmim2022} & 800 & 83.8 & - \\
SemMAE~\cite{semmae2022} & 800 & 83.4 & - \\
LocalMIM~\cite{localmim2023} & 1600 & 84.0 & - \\
HPM~\cite{hpm2023} & 800 & 84.2 & 85.8 \\
\midrule
\multicolumn{4}{l}{\textit{Masked Image Modeling + Contrastive Learning}} \\
iBOT~\cite{ibot2021} & 1600 & 84.0 & - \\
BootMAE~\cite{bootmae2022} & 800 & 84.2 & 85.9 \\
\midrule
\multicolumn{4}{l}{\textit{Autoregression Image Modeling}} \\
iGPT~\cite{igpt2020} & - & 72.6$^{\ddag}$ & - \\
ViT-iGPT~\cite{igpt2020} & - & 82.7$^{\ddag}$ & - \\
RandSAC~\cite{randsac2022} & 1600 & 83.9 & - \\
SAIM~\cite{saim2022} & 800 & 83.9 & - \\
\textbf{\method} & 400 & 83.8 & 85.5 \\
\textbf{\method} & 800 & 84.1 & 85.8 \\
\textbf{\method}$^\dag$ & 800 & \textbf{85.3} & \textbf{86.5} \\
\bottomrule
\end{tabular}
\caption{
\textbf{Comparison with previous methods on ImageNet-1K classification.}
All methods are evaluated by fine-tuning.
The resolution of images is 224$\times$224 for both pre-training and fine-tuning.
$\dag$ means using CLIP~\cite{clip_2021} feature as predict targets.
$\ddag$ means the result is borrowed from~\cite{saim2022}.
}
\label{tab:in1k}
\end{table}

\noindent\textbf{COCO object detection and segmentation.}
Following the configuration of ViTDet~\cite{vitdet2022}, the pre-trained models are fine-tuned on COCO~\cite{coco} with 100 epochs, using the Mask R-CNN~\cite{maskrcnn2017} detector.
We take ViT-B/16~\cite{vit2020} as the backbone.
AP$^{\text{b}}$ and AP$^{\text{m}}$ are adopted as the metric for object detection and instance segmentation, respectively.
As shown in \cref{tab:det_seg}, with only 400 epochs of pre-training, \method achieves 50.7\% AP$^{\text{b}}$ and 45.0\% AP$^{\text{m}}$, outperforming the baseline SAIM~\cite{saim2022} pre-trained for 800 epochs by +1.3\% and +1.0\%, respectively.
With 800 epochs of pre-training, \method achieves 51.3\% AP$^{\text{b}}$ and 45.4\% AP$^{\text{m}}$, surpassing SAIM~\cite{saim2022} by +1.9\% and +1.4\%, respectively.
The improvement is more significant than those on the ImageNet classification shown in \cref{tab:in1k}, indicating that semantic-aware autoregression modeling can learn better spatial reasoning abilities.

\begin{table}[t]
\centering
\setlength{\tabcolsep}{1.5pt}
\scalebox{0.92}{
\begin{tabular}{llclll}
\toprule
\multirow{2}{*}{method} & \multirow{2}{*}{ep.} & \multicolumn{2}{c}{COCO} & ADE20k \\
& & AP$^{\text{b}}$ & AP$^{\text{m}}$ & mIoU \\
\midrule
supervised & & 47.9 & 42.9 & 47.4 \\
MoCo v3~\cite{mocov3_2021} & 600 & 47.9$^{\dag}$ & 42.7$^{\dag}$ & 47.3$^{\dag}$ \\
BEiT~\cite{beit2021} & 800 & 49.8$^{\dag}$ & 44.4$^{\dag}$ & 47.1 \\
MAE~\cite{mae2022} & 1600 & 50.3 & 44.9 & \textbf{48.1} \\
CAE~\cite{cae2022} & 1600 & 50.1 & 44.0 & 48.8 \\
PeCo~\cite{peco2022} & 1600 & 49.1 & 43.8 & 48.5 \\
SIM~\cite{sim2023} & 800 & 49.1 & 43.8 & - \\
HPM~\cite{hpm2023} &  800 & 50.1 & 44.6 & 48.5 \\
SAIM~\cite{saim2022} & 800 & 49.4 & 44.0 & 47.8 \\
%
\textbf{\method} & 400 & 50.7 & 45.0  & 47.7 \\
\textbf{\method} & 800 & \textbf{51.3} & \textbf{45.4} & 48.0 \\
\bottomrule
\end{tabular}}
\caption{
\textbf{Comparison with other methods on downstream tasks.}
All methods take the ViT-B/16~\cite{vit2020} as the backbone.
For COCO~\cite{coco} object detection and instance segmentation, we utilize Mask R-CNN~\cite{maskrcnn2017} and perform 100 epoch fine-tuning.
For ADE20k~\cite{ade20k} semantic segmentation, we use UperNet~\cite{upernet2018} and perform 160k iterations of fine-tuning.
$\dag$ means the result is borrowed from~\cite{mae2022}.
}
\label{tab:det_seg}
\end{table}

\noindent\textbf{ADE20k semantic segmentation.}
The pre-trained models are fine-tuned on ADE20k~\cite{ade20k}  for semantic segmentation with 160k iterations using UperNet~\cite{upernet2018}
We take ViT-B/16~\cite{vit2020} as the backbone and search for the optimal learning rate for each model.
mIoU is used as the metric.
As shown in \cref{tab:det_seg}, with 800 epochs of pre-training, \method achieves 48.0\% mIoU, surpassing the baseline SAIM~\cite{saim2022} by +0.2\%.

\section{Conclusion}
\label{sec:conclusion}

In this study, we present a semantic-aware autoregressive image modeling (\method) method for visual representation learning.
The key insight of \method is to model images from the most semantic patches to the less semantic patches autoregressively.
In \method, we first calculate a semantic-aware permutation of patches according to their feature similarities and then perform the autoregression procedure based on the permutation.
In addition, considering that the raw pixels of patches are low-level signals and are not ideal prediction targets for learning high-level semantic representation, we also explore utilizing the patch features as the prediction targets.
We conducted extensive experiments on a broad range of downstream tasks, including image classification, object detection, and instance/semantic segmentation, to evaluate the performance of \method.
The results demonstrate \method achieves state-of-the-art performance compared with other self-supervised methods.
This study demonstrates that it is crucial for autoregressive image modeling to perform a suitable, semantic-aware autoregressive permutation.
\method shows good performance for vision pre-training, and it is a unified pre-training task compared with language modeling in NLP. 
One potential limitation of \method is that we only consider the case of one center patch in the permutation generation process, which may influence the performance of images with multiple objects.
This limitation can be overcome by calculating multiple center patches in the permutation generation process.



\bibliography{aaai24}

\begin{thebibliography}{58}
\providecommand{\natexlab}[1]{#1}

\bibitem[{Bao, Dong, and Wei(2022)}]{beit2021}
Bao, H.; Dong, L.; and Wei, F. 2022.
\newblock Beit: Bert pre-training of image transformers.
\newblock In \emph{International Conference on Learning Representations
  (ICLR)}.

\bibitem[{Brown et~al.(2020)Brown, Mann, Ryder, Subbiah, Kaplan, Dhariwal,
  Neelakantan, Shyam, Sastry, Askell et~al.}]{gpt3_2020}
Brown, T.; Mann, B.; Ryder, N.; Subbiah, M.; Kaplan, J.~D.; Dhariwal, P.;
  Neelakantan, A.; Shyam, P.; Sastry, G.; Askell, A.; et~al. 2020.
\newblock Language models are few-shot learners.
\newblock \emph{Advances in neural information processing systems}, 33:
  1877--1901.

\bibitem[{Caron et~al.(2020)Caron, Misra, Mairal, Goyal, Bojanowski, and
  Joulin}]{swav_2020}
Caron, M.; Misra, I.; Mairal, J.; Goyal, P.; Bojanowski, P.; and Joulin, A.
  2020.
\newblock Unsupervised learning of visual features by contrasting cluster
  assignments.
\newblock \emph{Advances in Neural Information Processing Systems (NeurIPS)},
  33: 9912--9924.

\bibitem[{Caron et~al.(2021)Caron, Touvron, Misra, J{\'e}gou, Mairal,
  Bojanowski, and Joulin}]{dino_2021}
Caron, M.; Touvron, H.; Misra, I.; J{\'e}gou, H.; Mairal, J.; Bojanowski, P.;
  and Joulin, A. 2021.
\newblock Emerging properties in self-supervised vision transformers.
\newblock In \emph{Proceedings of the IEEE/CVF International Conference on
  Computer Vision (ICCV)}, 9650--9660.

\bibitem[{Chen et~al.(2020{\natexlab{a}})Chen, Radford, Child, Wu, Jun, Luan,
  and Sutskever}]{igpt2020}
Chen, M.; Radford, A.; Child, R.; Wu, J.; Jun, H.; Luan, D.; and Sutskever, I.
  2020{\natexlab{a}}.
\newblock Generative pretraining from pixels.
\newblock In \emph{International conference on machine learning}, 1691--1703.
  PMLR.

\bibitem[{Chen et~al.(2020{\natexlab{b}})Chen, Kornblith, Norouzi, and
  Hinton}]{simclr_2020}
Chen, T.; Kornblith, S.; Norouzi, M.; and Hinton, G. 2020{\natexlab{b}}.
\newblock A simple framework for contrastive learning of visual
  representations.
\newblock In \emph{International conference on machine learning}, 1597--1607.
  PMLR.

\bibitem[{Chen et~al.(2021)Chen, Saxena, Li, Fleet, and
  Hinton}]{chen2021pix2seq}
Chen, T.; Saxena, S.; Li, L.; Fleet, D.~J.; and Hinton, G. 2021.
\newblock Pix2seq: A language modeling framework for object detection.
\newblock \emph{arXiv preprint arXiv:2109.10852}.

\bibitem[{Chen et~al.(2022)Chen, Ding, Wang, Xin, Mo, Wang, Han, Luo, Zeng, and
  Wang}]{cae2022}
Chen, X.; Ding, M.; Wang, X.; Xin, Y.; Mo, S.; Wang, Y.; Han, S.; Luo, P.;
  Zeng, G.; and Wang, J. 2022.
\newblock Context autoencoder for self-supervised representation learning.
\newblock \emph{arXiv preprint arXiv:2202.03026}.

\bibitem[{Chen and He(2021)}]{simsiam_2021}
Chen, X.; and He, K. 2021.
\newblock Exploring simple siamese representation learning.
\newblock In \emph{Proceedings of the IEEE/CVF Conference on Computer Vision
  and Pattern Recognition (CVPR)}, 15750--15758.

\bibitem[{Chen, Xie, and He(2021)}]{mocov3_2021}
Chen, X.; Xie, S.; and He, K. 2021.
\newblock An empirical study of training self-supervised vision transformers.
\newblock In \emph{Proceedings of the IEEE/CVF International Conference on
  Computer Vision (ICCV)}, 9640--9649.

\bibitem[{Contributors(2020)}]{mmseg2020}
Contributors, M. 2020.
\newblock {MMSegmentation}: OpenMMLab Semantic Segmentation Toolbox and
  Benchmark.
\newblock \url{https://github.com/open-mmlab/mmsegmentation}.

\bibitem[{Devlin et~al.(2018)Devlin, Chang, Lee, and Toutanova}]{bert2018}
Devlin, J.; Chang, M.-W.; Lee, K.; and Toutanova, K. 2018.
\newblock Bert: Pre-training of deep bidirectional transformers for language
  understanding.
\newblock \emph{arXiv preprint arXiv:1810.04805}.

\bibitem[{Dong et~al.(2019)Dong, Yang, Wang, Wei, Liu, Wang, Gao, Zhou, and
  Hon}]{dong2019unified}
Dong, L.; Yang, N.; Wang, W.; Wei, F.; Liu, X.; Wang, Y.; Gao, J.; Zhou, M.;
  and Hon, H.-W. 2019.
\newblock Unified language model pre-training for natural language
  understanding and generation.
\newblock \emph{Advances in neural information processing systems}, 32.

\bibitem[{Dong et~al.(2022)Dong, Bao, Zhang, Chen, Zhang, Yuan, Chen, Wen, and
  Yu}]{bootmae2022}
Dong, X.; Bao, J.; Zhang, T.; Chen, D.; Zhang, W.; Yuan, L.; Chen, D.; Wen, F.;
  and Yu, N. 2022.
\newblock Bootstrapped Masked Autoencoders for Vision BERT Pretraining.
\newblock In \emph{European Conference on Computer Vision (ECCV)}, 247--264.
  Springer.

\bibitem[{Dong et~al.(2023)Dong, Bao, Zhang, Chen, Zhang, Yuan, Chen, Wen, and
  Yu}]{peco2022}
Dong, X.; Bao, J.; Zhang, T.; Chen, D.; Zhang, W.; Yuan, L.; Chen, D.; Wen, F.;
  and Yu, N. 2023.
\newblock Peco: Perceptual codebook for bert pre-training of vision
  transformers.
\newblock In \emph{Proceedings of the AAAI Conference on Artificial
  Intelligence (AAAI)}.

\bibitem[{Dosovitskiy et~al.(2020)Dosovitskiy, Beyer, Kolesnikov, Weissenborn,
  Zhai, Unterthiner, Dehghani, Minderer, Heigold, Gelly et~al.}]{vit2020}
Dosovitskiy, A.; Beyer, L.; Kolesnikov, A.; Weissenborn, D.; Zhai, X.;
  Unterthiner, T.; Dehghani, M.; Minderer, M.; Heigold, G.; Gelly, S.; et~al.
  2020.
\newblock An image is worth 16x16 words: Transformers for image recognition at
  scale.
\newblock \emph{arXiv preprint arXiv:2010.11929}.

\bibitem[{Eriksen and Hoffman(1972)}]{1972_humanattention}
Eriksen, C.~W.; and Hoffman, J.~E. 1972.
\newblock Temporal and spatial characteristics of selective encoding from
  visual displays.
\newblock \emph{Perception \& psychophysics}, 12: 201--204.

\bibitem[{Everingham et~al.(2009)Everingham, Van~Gool, Williams, Winn, and
  Zisserman}]{everingham2009pascal}
Everingham, M.; Van~Gool, L.; Williams, C.~K.; Winn, J.; and Zisserman, A.
  2009.
\newblock The pascal visual object classes (voc) challenge.
\newblock \emph{International Journal of Computer Vision (IJCV)}, 88: 303--308.

\bibitem[{Guo et~al.(2022)Guo, Xu, Li, Ni, Zhu, Sun, and Xu}]{hcsc_2022}
Guo, Y.; Xu, M.; Li, J.; Ni, B.; Zhu, X.; Sun, Z.; and Xu, Y. 2022.
\newblock HCSC: Hierarchical Contrastive Selective Coding.
\newblock In \emph{Proceedings of the IEEE/CVF Conference on Computer Vision
  and Pattern Recognition (CVPR)}, 9706--9715.

\bibitem[{He et~al.(2022)He, Chen, Xie, Li, Doll{\'a}r, and Girshick}]{mae2022}
He, K.; Chen, X.; Xie, S.; Li, Y.; Doll{\'a}r, P.; and Girshick, R. 2022.
\newblock Masked autoencoders are scalable vision learners.
\newblock In \emph{Proceedings of the IEEE/CVF Conference on Computer Vision
  and Pattern Recognition}, 16000--16009.

\bibitem[{He et~al.(2020)He, Fan, Wu, Xie, and Girshick}]{moco_2020}
He, K.; Fan, H.; Wu, Y.; Xie, S.; and Girshick, R. 2020.
\newblock Momentum contrast for unsupervised visual representation learning.
\newblock In \emph{Proceedings of the IEEE/CVF Conference on Computer Vision
  and Pattern Recognition (CVPR)}, 9729--9738.

\bibitem[{He et~al.(2017)He, Gkioxari, Doll{\'a}r, and Girshick}]{maskrcnn2017}
He, K.; Gkioxari, G.; Doll{\'a}r, P.; and Girshick, R. 2017.
\newblock Mask r-cnn.
\newblock In \emph{Proceedings of the IEEE/CVF International Conference on
  Computer Vision (ICCV)}, 2961--2969.

\bibitem[{Hou et~al.(2023)Hou, Sun, Chen, Xie, and Kung}]{milan2022}
Hou, Z.; Sun, F.; Chen, Y.-K.; Xie, Y.; and Kung, S.-Y. 2023.
\newblock Milan: Masked image pretraining on language assisted representation.

\bibitem[{Hua et~al.(2022)Hua, Tian, Ren, Zhao, and Sigal}]{randsac2022}
Hua, T.; Tian, Y.; Ren, S.; Zhao, H.; and Sigal, L. 2022.
\newblock Self-supervision through random segments with autoregressive coding
  (randsac).
\newblock \emph{International Conference on Learning Representations (ICLR)}.

\bibitem[{Kakogeorgiou et~al.(2022)Kakogeorgiou, Gidaris, Psomas, Avrithis,
  Bursuc, Karantzalos, and Komodakis}]{attmask2022}
Kakogeorgiou, I.; Gidaris, S.; Psomas, B.; Avrithis, Y.; Bursuc, A.;
  Karantzalos, K.; and Komodakis, N. 2022.
\newblock What to hide from your students: Attention-guided masked image
  modeling.
\newblock In \emph{Computer Vision--ECCV 2022: 17th European Conference, Tel
  Aviv, Israel, October 23--27, 2022, Proceedings, Part XXX}, 300--318.
  Springer.

\bibitem[{Koch et~al.(2006)Koch, McLean, Segev, Freed, Berry, Balasubramanian,
  and Sterling}]{2006_humanattention}
Koch, K.; McLean, J.; Segev, R.; Freed, M.~A.; Berry, M.~J.; Balasubramanian,
  V.; and Sterling, P. 2006.
\newblock How much the eye tells the brain.
\newblock \emph{Current biology}, 16(14): 1428--1434.

\bibitem[{Komodakis and Gidaris(2018)}]{komodakis2018unsupervised}
Komodakis, N.; and Gidaris, S. 2018.
\newblock Unsupervised representation learning by predicting image rotations.
\newblock In \emph{International conference on learning representations
  (ICLR)}.

\bibitem[{Li et~al.(2022{\natexlab{a}})Li, Zheng, Liu, Wang, Su, and
  Zheng}]{semmae2022}
Li, G.; Zheng, H.; Liu, D.; Wang, C.; Su, B.; and Zheng, C. 2022{\natexlab{a}}.
\newblock SemMAE: Semantic-Guided Masking for Learning Masked Autoencoders.
\newblock In \emph{Advances in Neural Information Processing Systems
  (NeurIPS)}.

\bibitem[{Li et~al.(2022{\natexlab{b}})Li, Mao, Girshick, and He}]{vitdet2022}
Li, Y.; Mao, H.; Girshick, R.; and He, K. 2022{\natexlab{b}}.
\newblock Exploring plain vision transformer backbones for object detection.
\newblock In \emph{European Conference on Computer Vision (ECCV)}, 280--296.
  Springer.

\bibitem[{Lin et~al.(2017)Lin, Doll{\'a}r, Girshick, He, Hariharan, and
  Belongie}]{fpn2017}
Lin, T.-Y.; Doll{\'a}r, P.; Girshick, R.; He, K.; Hariharan, B.; and Belongie,
  S. 2017.
\newblock Feature pyramid networks for object detection.
\newblock In \emph{Proceedings of the IEEE/CVF Conference on Computer Vision
  and Pattern Recognition (CVPR)}, 2117--2125.

\bibitem[{Lin et~al.(2014)Lin, Maire, Belongie, Hays, Perona, Ramanan,
  Doll{\'a}r, and Zitnick}]{coco}
Lin, T.-Y.; Maire, M.; Belongie, S.; Hays, J.; Perona, P.; Ramanan, D.;
  Doll{\'a}r, P.; and Zitnick, C.~L. 2014.
\newblock Microsoft coco: Common objects in context.
\newblock In \emph{European Conference on Computer Vision (ECCV)}, 740--755.
  Springer.

\bibitem[{Noroozi and Favaro(2016)}]{noroozi2016unsupervised}
Noroozi, M.; and Favaro, P. 2016.
\newblock Unsupervised learning of visual representations by solving jigsaw
  puzzles.
\newblock In \emph{Computer Vision--ECCV 2016: 14th European Conference,
  Amsterdam, The Netherlands, October 11-14, 2016, Proceedings, Part VI},
  69--84. Springer.

\bibitem[{Noroozi, Pirsiavash, and Favaro(2017)}]{noroozi2017representation}
Noroozi, M.; Pirsiavash, H.; and Favaro, P. 2017.
\newblock Representation learning by learning to count.
\newblock In \emph{Proceedings of the IEEE international conference on computer
  vision}, 5898--5906.

\bibitem[{Oord, Li, and Vinyals(2018)}]{cpc2018}
Oord, A. v.~d.; Li, Y.; and Vinyals, O. 2018.
\newblock Representation learning with contrastive predictive coding.
\newblock \emph{arXiv preprint arXiv:1807.03748}.

\bibitem[{Pathak et~al.(2016)Pathak, Krahenbuhl, Donahue, Darrell, and
  Efros}]{pathak2016context}
Pathak, D.; Krahenbuhl, P.; Donahue, J.; Darrell, T.; and Efros, A.~A. 2016.
\newblock Context encoders: Feature learning by inpainting.
\newblock In \emph{Proceedings of the IEEE conference on computer vision and
  pattern recognition}, 2536--2544.

\bibitem[{Qi et~al.(2022)Qi, Yang, Zhu, Liu, Wu, Zhao, and Li}]{saim2022}
Qi, Y.; Yang, F.; Zhu, Y.; Liu, Y.; Wu, L.; Zhao, R.; and Li, W. 2022.
\newblock Exploring Stochastic Autoregressive Image Modeling for Visual
  Representation.
\newblock \emph{Proceedings of the AAAI Conference on Artificial Intelligence
  (AAAI)}.

\bibitem[{Radford et~al.(2021)Radford, Kim, Hallacy, Ramesh, Goh, Agarwal,
  Sastry, Askell, Mishkin, Clark et~al.}]{clip_2021}
Radford, A.; Kim, J.~W.; Hallacy, C.; Ramesh, A.; Goh, G.; Agarwal, S.; Sastry,
  G.; Askell, A.; Mishkin, P.; Clark, J.; et~al. 2021.
\newblock Learning transferable visual models from natural language
  supervision.
\newblock In \emph{International Conference on Machine Learning (ICML)},
  8748--8763. PMLR.

\bibitem[{Radford et~al.(2018)Radford, Narasimhan, Salimans, Sutskever
  et~al.}]{gpt1_2018}
Radford, A.; Narasimhan, K.; Salimans, T.; Sutskever, I.; et~al. 2018.
\newblock Improving language understanding by generative pre-training.

\bibitem[{Radford et~al.(2019)Radford, Wu, Child, Luan, Amodei, Sutskever
  et~al.}]{gpt2_2019}
Radford, A.; Wu, J.; Child, R.; Luan, D.; Amodei, D.; Sutskever, I.; et~al.
  2019.
\newblock Language models are unsupervised multitask learners.
\newblock \emph{OpenAI blog}, 1(8): 9.

\bibitem[{Ramesh et~al.(2021)Ramesh, Pavlov, Goh, Gray, Voss, Radford, Chen,
  and Sutskever}]{ramesh2021zero}
Ramesh, A.; Pavlov, M.; Goh, G.; Gray, S.; Voss, C.; Radford, A.; Chen, M.; and
  Sutskever, I. 2021.
\newblock Zero-shot text-to-image generation.
\newblock In \emph{International Conference on Machine Learning}, 8821--8831.
  PMLR.

\bibitem[{Russakovsky et~al.(2015)Russakovsky, Deng, Su, Krause, Satheesh, Ma,
  Huang, Karpathy, Khosla, Bernstein et~al.}]{imagenet}
Russakovsky, O.; Deng, J.; Su, H.; Krause, J.; Satheesh, S.; Ma, S.; Huang, Z.;
  Karpathy, A.; Khosla, A.; Bernstein, M.; et~al. 2015.
\newblock Imagenet large scale visual recognition challenge.
\newblock \emph{International Journal of Computer Vision (IJCV)}, 115:
  211--252.

\bibitem[{Song et~al.(2020)Song, Tan, Qin, Lu, and Liu}]{song2020mpnet}
Song, K.; Tan, X.; Qin, T.; Lu, J.; and Liu, T.-Y. 2020.
\newblock Mpnet: Masked and permuted pre-training for language understanding.
\newblock \emph{Advances in Neural Information Processing Systems}, 33:
  16857--16867.

\bibitem[{Song et~al.(2023{\natexlab{a}})Song, Xie, Zhang, and Luo}]{mokd_2023}
Song, K.; Xie, J.; Zhang, S.; and Luo, Z. 2023{\natexlab{a}}.
\newblock Multi-Mode Online Knowledge Distillation for Self-Supervised Visual
  Representation Learning.
\newblock In \emph{Proceedings of the IEEE/CVF Conference on Computer Vision
  and Pattern Recognition (CVPR)}, 11848--11857.

\bibitem[{Song et~al.(2023{\natexlab{b}})Song, Zhang, Luo, Wang, and
  Xie}]{scfs_2023}
Song, K.; Zhang, S.; Luo, Z.; Wang, T.; and Xie, J. 2023{\natexlab{b}}.
\newblock Semantics-Consistent Feature Search for Self-Supervised Visual
  Representation Learning.
\newblock In \emph{Proceedings of the IEEE/CVF International Conference on
  Computer Vision (ICCV)}, 16099--16108.

\bibitem[{Tao et~al.(2023)Tao, Zhu, Su, Huang, Li, Zhou, Qiao, Wang, and
  Dai}]{sim2023}
Tao, C.; Zhu, X.; Su, W.; Huang, G.; Li, B.; Zhou, J.; Qiao, Y.; Wang, X.; and
  Dai, J. 2023.
\newblock Siamese image modeling for self-supervised vision representation
  learning.
\newblock In \emph{Proceedings of the IEEE/CVF Conference on Computer Vision
  and Pattern Recognition}, 2132--2141.

\bibitem[{Van~den Oord et~al.(2016)Van~den Oord, Kalchbrenner, Espeholt,
  Vinyals, Graves et~al.}]{van2016conditional}
Van~den Oord, A.; Kalchbrenner, N.; Espeholt, L.; Vinyals, O.; Graves, A.;
  et~al. 2016.
\newblock Conditional image generation with pixelcnn decoders.
\newblock \emph{Advances in neural information processing systems}, 29.

\bibitem[{Vaswani et~al.(2017)Vaswani, Shazeer, Parmar, Uszkoreit, Jones,
  Gomez, Kaiser, and Polosukhin}]{transformer2017}
Vaswani, A.; Shazeer, N.; Parmar, N.; Uszkoreit, J.; Jones, L.; Gomez, A.~N.;
  Kaiser, {\L}.; and Polosukhin, I. 2017.
\newblock Attention is all you need.
\newblock \emph{Advances in Neural Information Processing Systems (NeurIPS)},
  30.

\bibitem[{Wang et~al.(2023{\natexlab{a}})Wang, Song, Fan, Wang, Xie, and
  Zhang}]{hpm2023}
Wang, H.; Song, K.; Fan, J.; Wang, Y.; Xie, J.; and Zhang, Z.
  2023{\natexlab{a}}.
\newblock Hard Patches Mining for Masked Image Modeling.
\newblock In \emph{Proceedings of the IEEE/CVF Conference on Computer Vision
  and Pattern Recognition (CVPR)}.

\bibitem[{Wang et~al.(2023{\natexlab{b}})Wang, Tang, Wang, Guo, Deng, and
  Han}]{localmim2023}
Wang, H.; Tang, Y.; Wang, Y.; Guo, J.; Deng, Z.-H.; and Han, K.
  2023{\natexlab{b}}.
\newblock Masked Image Modeling with Local Multi-Scale Reconstruction.
\newblock In \emph{Proceedings of the IEEE/CVF Conference on Computer Vision
  and Pattern Recognition (CVPR)}.

\bibitem[{Wei et~al.(2022)Wei, Fan, Xie, Wu, Yuille, and
  Feichtenhofer}]{maskfeat2022}
Wei, C.; Fan, H.; Xie, S.; Wu, C.-Y.; Yuille, A.; and Feichtenhofer, C. 2022.
\newblock Masked feature prediction for self-supervised visual pre-training.
\newblock In \emph{Proceedings of the IEEE/CVF Conference on Computer Vision
  and Pattern Recognition}, 14668--14678.

\bibitem[{Wu et~al.(2019)Wu, Kirillov, Massa, Lo, and Girshick}]{detectron2}
Wu, Y.; Kirillov, A.; Massa, F.; Lo, W.-Y.; and Girshick, R. 2019.
\newblock Detectron2.
\newblock \url{https://github.com/facebookresearch/detectron2}.

\bibitem[{Wu et~al.(2018)Wu, Xiong, Yu, and Lin}]{insdis_2018}
Wu, Z.; Xiong, Y.; Yu, S.~X.; and Lin, D. 2018.
\newblock Unsupervised feature learning via non-parametric instance
  discrimination.
\newblock In \emph{Proceedings of the IEEE/CVF Conference on Computer Vision
  and Pattern Recognition (CVPR)}, 3733--3742.

\bibitem[{Xiao et~al.(2018)Xiao, Liu, Zhou, Jiang, and Sun}]{upernet2018}
Xiao, T.; Liu, Y.; Zhou, B.; Jiang, Y.; and Sun, J. 2018.
\newblock Unified perceptual parsing for scene understanding.
\newblock In \emph{European Conference on Computer Vision (ECCV)}, 418--434.

\bibitem[{Xie et~al.(2022)Xie, Zhang, Cao, Lin, Bao, Yao, Dai, and
  Hu}]{simmim2022}
Xie, Z.; Zhang, Z.; Cao, Y.; Lin, Y.; Bao, J.; Yao, Z.; Dai, Q.; and Hu, H.
  2022.
\newblock Simmim: A simple framework for masked image modeling.
\newblock In \emph{Proceedings of the IEEE/CVF Conference on Computer Vision
  and Pattern Recognition}, 9653--9663.

\bibitem[{Yang et~al.(2019)Yang, Dai, Yang, Carbonell, Salakhutdinov, and
  Le}]{yang2019xlnet}
Yang, Z.; Dai, Z.; Yang, Y.; Carbonell, J.; Salakhutdinov, R.~R.; and Le, Q.~V.
  2019.
\newblock Xlnet: Generalized autoregressive pretraining for language
  understanding.
\newblock \emph{Advances in neural information processing systems}, 32.

\bibitem[{Zhang, Isola, and Efros(2016)}]{zhang2016colorful}
Zhang, R.; Isola, P.; and Efros, A.~A. 2016.
\newblock Colorful image colorization.
\newblock In \emph{Computer Vision--ECCV 2016: 14th European Conference,
  Amsterdam, The Netherlands, October 11-14, 2016, Proceedings, Part III 14},
  649--666. Springer.

\bibitem[{Zhou et~al.(2017)Zhou, Zhao, Puig, Fidler, Barriuso, and
  Torralba}]{ade20k}
Zhou, B.; Zhao, H.; Puig, X.; Fidler, S.; Barriuso, A.; and Torralba, A. 2017.
\newblock Scene parsing through ade20k dataset.
\newblock In \emph{Proceedings of the IEEE/CVF Conference on Computer Vision
  and Pattern Recognition (CVPR)}, 633--641.

\bibitem[{Zhou et~al.(2022)Zhou, Wei, Wang, Shen, Xie, Yuille, and
  Kong}]{ibot2021}
Zhou, J.; Wei, C.; Wang, H.; Shen, W.; Xie, C.; Yuille, A.; and Kong, T. 2022.
\newblock Image BERT Pre-training with Online Tokenizer.
\newblock In \emph{International Conference on Learning Representations
  (ICLR)}.

\end{thebibliography}



\newpage

\renewcommand\thefigure{S\arabic{figure}}
\renewcommand\thetable{S\arabic{table}}
\renewcommand\theequation{S\arabic{equation}}
\setcounter{equation}{0}
\setcounter{table}{0}
\setcounter{figure}{0}
\setcounter{section}{0}
\renewcommand\thesection{\Alph{section}}

\newpage
\clearpage

\section*{Appendix}


\subsection{Implementation details}
\label{sec_app_detail}

\noindent\textbf{ViT architecture.}
We follow the standard vanilla ViT~\cite{vit2020} architecture as the backbone, which is a stack of Transformer blocks~\cite{transformer2017}.
Following MAE~\cite{mae2022} and SAIM~\cite{saim2022}, we use the fixed 2D sine-cosine positional embeddings during pre-training.

\noindent\textbf{Hyperparameters for pre-training and fine-tuning on ImageNet.}
For all experiments in this paper, we take ImageNet-1K~\cite{imagenet} as the pre-training dataset.
Pre-training and fine-tuning details can be found in \cref{tab:pretrain_hyperparameters} and \cref{tab:finetune_hyperparameters}, respectively.
Most of the configurations are borrowed from SAIM~\cite{saim2022}.
The linear learning rate scaling rule is adopted: $lr = lr_{\mathrm{base}} \times \mathrm{batch\_size}\ /\ 256$.
For ViT-B/16, pre-training and fine-tuning are conducted with 32 and 16 Tesla V100 GPUs, respectively.
For ViT-L/16, pre-training and fine-tuning are conducted with 64 and 16 Tesla V100 GPUs, respectively.

\begin{table}[h]
\centering
\begin{tabular}{l|l}
\cline{1-2}
config                 & value                    \\ \cline{1-2}
optimizer              & AdamW                    \\
base learning rate     & 2e-4                   \\
weight decay           & 0.05                     \\
optimizer momentum     & $\beta_{1}$,  $\beta_{2}$=0.9, 0.95  \\
batch size             & 2048                     \\
learning rate schedule & cosine decay             \\
warmup epochs          & 30                       \\
augmentation           & RandomResizeCrop         \\ \cline{1-2}
\end{tabular}
\caption{Hyperparameters for pertaining on ImageNet-1k.}
\label{tab:pretrain_hyperparameters}
\end{table}

\begin{table}[h]
\centering
\begin{tabular}{l|l}
\hline
config                 & value                           \\ \hline
optimizer              & AdamW                           \\
base learning rate     & 5e-4                            \\
weight decay           & 0.05                            \\
optimizer momentum     & $\beta_1$,  $\beta_2$ = 0.9, 0.99 \\
layer-wise lr decay    & 0.65                            \\
batch size             & 1024                            \\
learning rate schedule & cosine decay                    \\
warmup epochs          & 20                               \\
training epoch         & 100                             \\
augmentation           & RandAug       (9, 0.5)                 \\
label smoothing        & 0.1                             \\
mixup                  & 0.8                             \\
cutmix                 & 1.0                             \\
drop path              & 0.1                             \\ \hline
\end{tabular}
\caption{Hyperparameters for finetuning on ImageNet-1k.}
\label{tab:finetune_hyperparameters}
\end{table}

\noindent\textbf{Hyperparameters for object detection and instance segmentation on COCO.}
We take Mask R-CNN~\cite{maskrcnn2017} with FPN~\cite{fpn2017} as the object detector.
Following~\cite{mae2022}, to obtain pyramid feature maps for matching the requirements of FPN~\cite{fpn2017}, we equally divide the backbone into 4 subsets, and then apply convolutions to get the intermediate feature maps at different scales (stride 4, 8, 16, or 32).
The hyperparameters used for finetuning \method on COCO~\cite{coco} are shown in the ~\cref{tab:object detection and instance segmentation on COCO}.
The batch size is 64, the warmup epoch is 0.25, and the weight decay is 0.1. For ViT-B/16, we train 100 epochs with a learning rate of 8e-5.
Experiments are conducted on 32 Tesla V100 GPUs.

\begin{table}[ht]
\centering
\begin{tabular}{l|l}
\hline
config                 & value                           \\ \hline
optimizer              & AdamW                           \\
learning rate          & 8e-5                            \\
weight decay           & 0.1                           \\
optimizer momentum     & $\beta_1$,  $\beta_2$ = 0.9, 0.99 \\
batch size             & 64                            \\
learning rate schedule & cosine decay                    \\
warmup epochs          & 0.25                               \\
training epoch         & 100                             \\
drop path              & 0.1                             \\
input resolution       & $1024\times1024$    \\
position embedding interpolate  & bilinear \\
\hline
\end{tabular}
\caption{Hyperparameters for object detection and instance segmentation on COCO.}
\label{tab:object detection and instance segmentation on COCO}
\end{table}

\noindent\textbf{Hyperparameters for semantic segmentation on ADE20K.}
We take UperNet~\cite{upernet2018} as the segmentation decoder following the code of~\cite{mmseg2020, hpm2023}.
The hyperparameters used for finetuning \method on ADE20K~\cite{ade20k} are shown in the ~\cref{tab:semantic segmentation on ADE20K}.
We use layer-wise learning rate decay, weight decay, and AdamW.
The batch size is 16, the warmup iteration is 1500, and the weight decay is 0.05.
For ViT-B/16, we train 160k iterations with a learning rate of 4e-4.
All experiments are conducted on 8 Tesla V100 GPUs.

\begin{table}[h]
\centering
\begin{tabular}{l|l}
\hline
config                 & value                           \\ \hline
optimizer              & AdamW                           \\
learning rate          & 4e-4                            \\
weight decay           & 0.05                        \\
layer-wise lr decay    & 0.65                            \\
optimizer momentum     & $\beta_1$,  $\beta_2$ = 0.9, 0.99 \\
batch size             & 16                            \\
learning rate schedule & poly decay                    \\
warmup iterations          & 1500                               \\
training iterations         & 160k                           \\
drop path              & 0.1                             \\
input resolution       & $512\times512$    \\
position embedding interpolate  & bilinear \\
\hline
\end{tabular}
\caption{Hyperparameters for semantic segmentation on ADE20K.}
\label{tab:semantic segmentation on ADE20K}
\end{table}

\begin{figure*}[!h]
\centering
\includegraphics[width=0.98\linewidth]{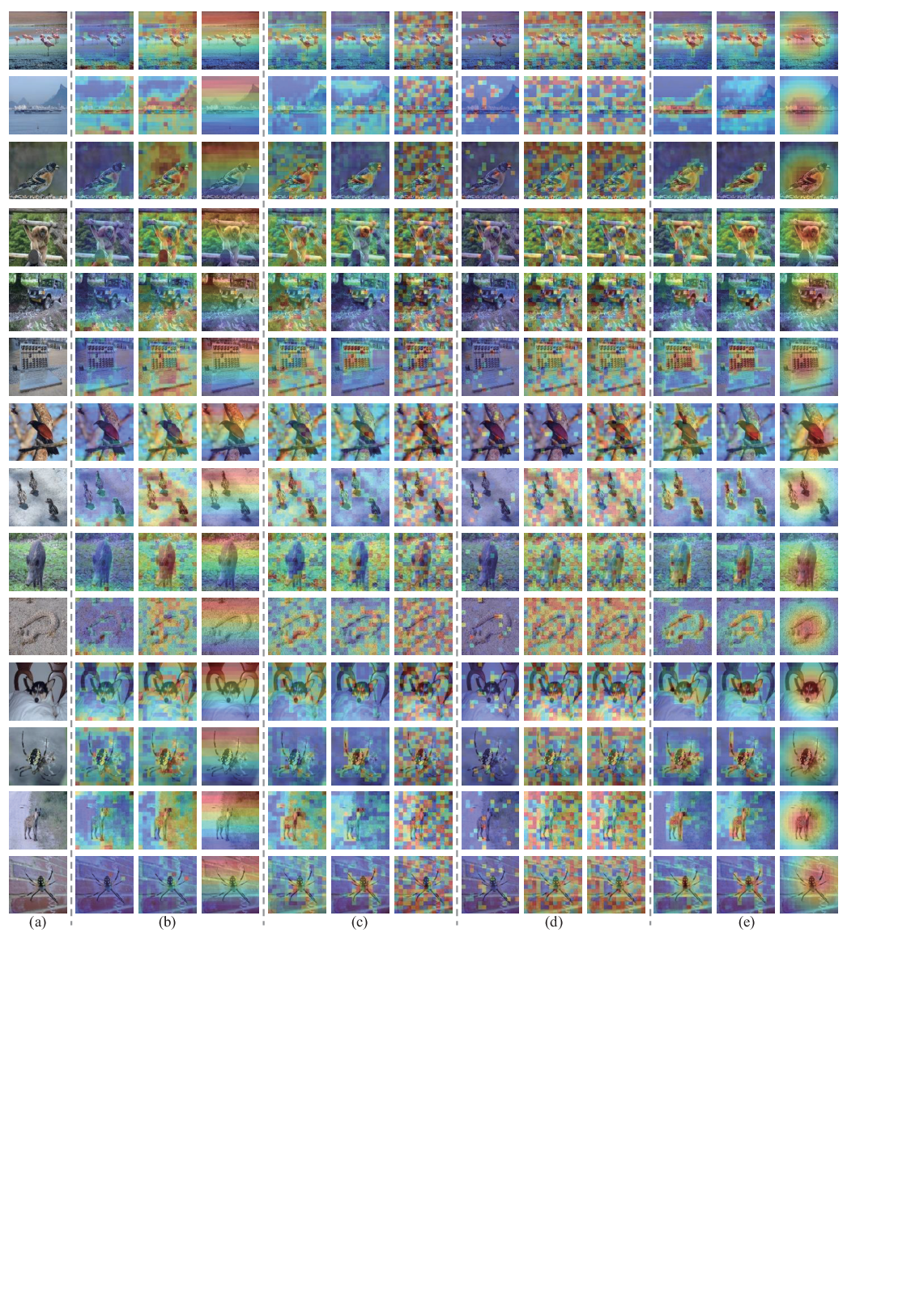}
\caption{
Visualization of different autoregression orders.
(a) input images, (b) raster order used in iGPT~\cite{igpt2020}, (c) stochastic order used in SAIM~\cite{saim2022}, (d) similarity order (the similarity map $S$ is also directly used as the autoregression order), and (e) semantic-aware order used in \method. In (b)(c)(d)(e), the first column shows the self-attention maps from the last block, the second column shows similarity maps $S$ from the last block, and the last column shows the corresponding autoregression orders (more warm-colored patches are predicted first).
}
\label{fig_app_visual}
\end{figure*}

\subsection{More Visualization Results}
\label{sec_app_vis}
We visualize the self-attention maps, the similarity maps $\vm{S}$, and the corresponding autoregression orders of each method in \cref{fig_app_visual}.
The self-attention maps and the similarity maps of the semantic-aware order used in \method locate on semantic regions more accurately than other methods.
This indicates that \method can learn more semantic representations.

\subsection{Pseudo-code}
\label{sec_app_code}
The training procedure of \method is summarized in \cref{alg:code}.

\begin{algorithm}[h]
\caption{Pseudo-Code of \method in a PyTorch-like Style.}
\label{alg:code}
\definecolor{codeblue}{rgb}{0.25,0.5,0.5}
\lstset{
  backgroundcolor=\color{white},
  basicstyle=\fontsize{7.2pt}{7.2pt}\ttfamily\selectfont,
  columns=fullflexible,
  breaklines=true,
  captionpos=b,
  commentstyle=\fontsize{7.2pt}{7.2pt}\color{codeblue},
  keywordstyle=\fontsize{7.2pt}{7.2pt},
}
\begin{lstlisting}[language=python]
# imgs: a minibatch with N samples
# targets: prediction targets
# pos_embed: fixed 2d sin_cos pos_embed
# depth: total number of encoder(decoder) blocks
# encoder: self-attention encoder_blocks
# decoder: cross-attention decoder_blocks and head

# initialize g and h
x = PatchEmbed(imgs)
h = x + pos_embed
g = pos_embed
# generate semantic-aware permutation
z = encoder(h)
C = permutation_generation(z) # Eq.(2)(3)(4)(5)
# generate attention mask
mask_h = C.unsqueeze(-1) >= C.unsqueeze(1)
mask_g = C.unsqueeze(-1) > C.unsqueeze(1)
# forward
for i in range(depth):
    h = encoder_blocks[i](q=h, kv=h, mask=mask_h)
    g = decoder_blocks[i](q=g, kv=h, mask=mask_g)
g = head(g) # MLP layer
# compute loss
loss = MSE(g, targets)
# backward
loss.backward()
update(encoder, decoder)
\end{lstlisting}
\end{algorithm}

\subsection{Comparison with CLIP backbones}
\label{sec_app_clip}
We compare the fine-tuned accuracy of CLIP backbones with \method on ImageNet.
The results in \cref{tab:compare_with_clip} show that \method can outperform the teacher models with AIM.

\begin{table}[h]
\centering
\begin{tabular}{l|cc}
\toprule
Method & ViT-B & ViT-L \\ \hline
CLIP & 82.1 & 85.3 \\ \hline
\method & 85.3 & 86.5 \\ \hline
\end{tabular}
\caption{Compared with CLIP~\cite{clip_2021} on ImageNet.}
\label{tab:compare_with_clip}
\end{table}

\end{document}